\documentclass[A4]{article}
\usepackage{graphicx}
\usepackage{amsmath,amssymb} 
\usepackage{color}

\DeclareMathOperator*{\argmax}{arg\,max}

\begin{document}
\pagestyle{headings}


\title{Visual place recognition using\\ landmark distribution descriptors} 

\author{Pilailuck Panphattarasap and Andrew Calway\\[1ex]Department of Computer Science\\University of Bristol, UK\\[1ex]{\tt \{pp12907,Andrew.Calway\}@bristol.ac.uk}}

\maketitle

\begin{abstract}
Recent work by S\"{u}nderhauf et al. \cite{Suenderhauf-RSS-15} demonstrated improved visual place recognition using proposal regions coupled with features from convolutional neural networks (CNN) to match landmarks between views. In this work we extend the approach by introducing descriptors built from landmark features which also encode the spatial distribution of the landmarks within a view. Matching descriptors then enforces consistency of the relative positions of landmarks between views. This has a significant impact on performance. For example, in experiments on 10 image-pair datasets, each consisting of 200 urban locations with significant differences in viewing positions and conditions, we recorded average precision of around 70\% (at 100\% recall), compared with 58\% obtained using whole image CNN features and 50\% for the method in \cite{Suenderhauf-RSS-15}. 
\end{abstract}
\section{Introduction}

Visual place recognition is the task of matching a view of a place with a different view of the same place taken at a different time. If incorporated into a mapping framework, such as a topological representation of places, for example, then reliable and fast visual place recognition opens up the possibility of truly autonomous navigation, with applications in robotics and related areas. It would negate the need for a positioning infrastructure such as GPS and perhaps more interestingly, in respect of human-robot interaction, be more akin to the wayfinding techniques employed by humans.

Automated recognition of places based on visual information is however very challenging. It is highly dependent on the characteristics of places, the viewing positions and directions, and the environmental conditions in terms of light and visibility. Perspective effects, occlusions, changes in natural vegetation, differences in seasonal and day/night appearance, and the presence of transient objects such as vehicles and people, all conspire to make recognition in its most general form a very hard problem.

Research into using computer vision to recognise places has made progress, as documented by Lowry et al. \cite{Lowry-TOR-2015}. Broadly speaking, approaches to date fall into two main categories: those based on matching local features between views; and those based on comparing whole image characteristics. Of the former, techniques based around the shift-invariant feature transform (SIFT) \cite{Lowe-IJCV-2004} and its variants are the most common, whilst in the latter category the GIST descriptor \cite{Oliva-IJCV-2001} has found widespread use. To aid robustness, these techniques are often incorporated within some form of temporal integration, the probabilistic FAB-MAP method \cite{Cummins-IJRR-2008,Cummins-IJRR-2010} being the most well known. Other techniques aim to deal with seasonal, day/night and long term changes \cite{Lowry-TOR-2015}. 

As pointed out in \cite{Lowry-TOR-2015}, the two categories above tend to address complementary issues: local features provide a degree of invariance to viewing position and direction, whilst global descriptors provide better invariance to changes in viewing conditions. However, neither do both. To address this, recent work, for example that described in \cite{McManus-RSS-2014} and \cite{Suenderhauf-RSS-15}, match local regions corresponding to salient landmarks in the scene such as buildings, trees, windows, etc. Matching these regions using global-type descriptors provides a degree of invariance to changing conditions, whilst their localised nature gives better invariance to viewing position and direction. We adopt a similar approach in this work.

\begin{figure}[t]
\begin{center}
\begin{tabular}{cc}
\begin{tabular}{cc}
\includegraphics[width=0.25\textwidth]{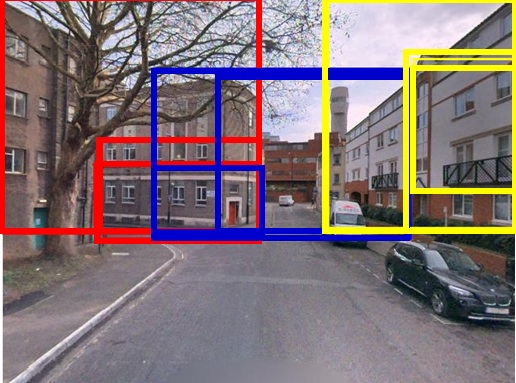}&
\includegraphics[width=0.25\textwidth]{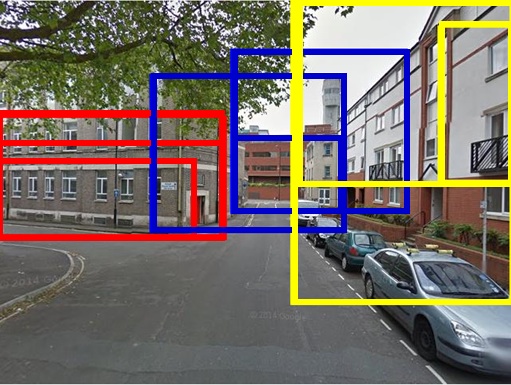}\\[3ex]
\multicolumn{2}{c}{\includegraphics[width=0.5\textwidth]{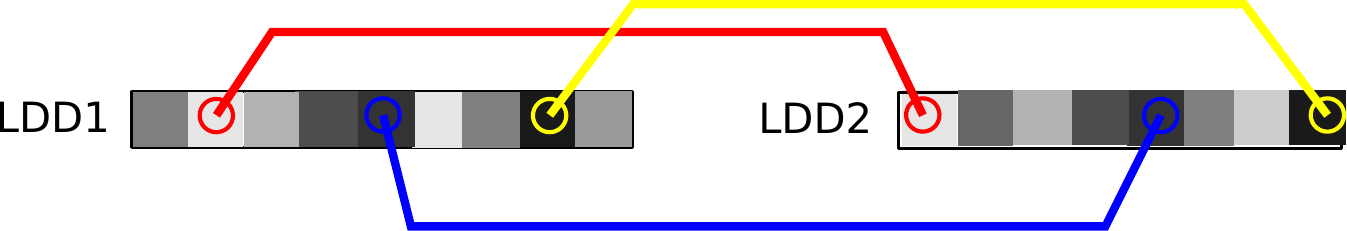}}\\[3ex]
\includegraphics[width=0.25\textwidth]{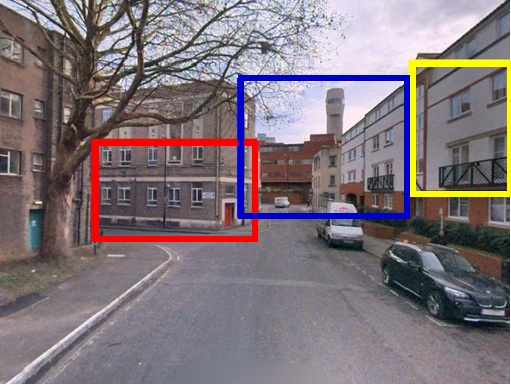}&
\includegraphics[width=0.25\textwidth]{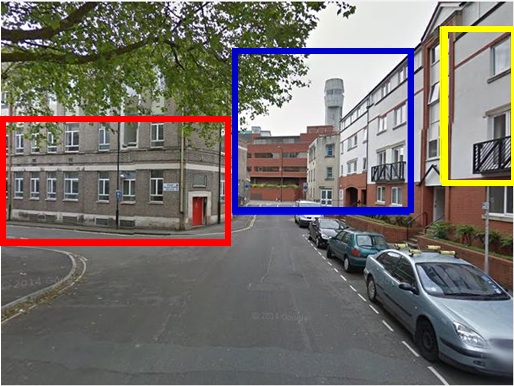}
\end{tabular}
&
\begin{tabular}{c}
\includegraphics[width=0.35\textwidth]{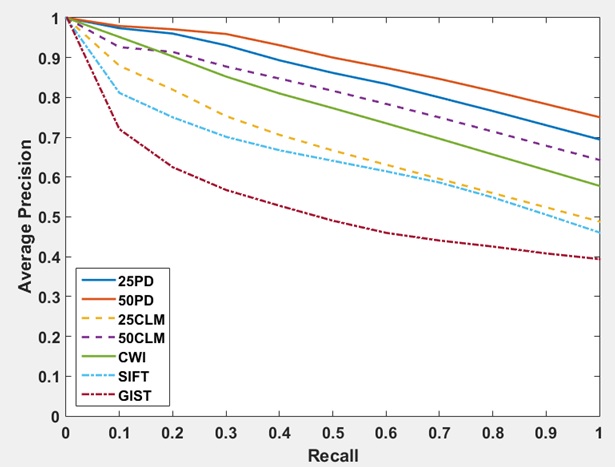}\\
\includegraphics[width=0.35\textwidth]{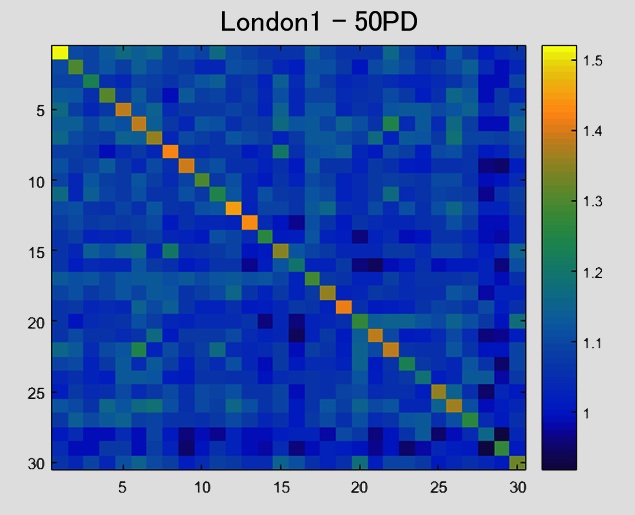}
\end{tabular}
\\[2ex]
(a)&(b)\\
\end{tabular}
\end{center}
\vspace*{-3ex}
\caption{Place recognition using landmark distribution descriptors (LDDs). Proposal regions from {\em Edge Boxes}  (top left) are represented by CNN feature vectors and stacked in horizontal position order into an LDD for the view (left middle). The similarity of top matching regions within sections of the descriptor are then used as a measure of similarity between the views (bottom left). The approach outperforms comparison methods over 10 datasets each with 200 urban locations (top right) and shows excellent discrimination characteristics as illustrated by the confusion matrix in the bottom right.} 
\label{fig:showcase}
\vspace{-3ex}
\end{figure}

\subsection{Landmark distribution descriptors}

Our main contribution is that in addition to matching landmark regions, we seek to maintain consistency of the {spatial distribution of landmarks} between views of a place. In doing so, we aim to reduce the impact of similar landmarks being present in different places - although individual landmarks may match, it is their relative positions across the view that characterises the place. In this work we limit ourselves to cases in which the different views of a place contain the same panorama but viewed from a different angle and distance, so that to a reasonable approximation the order of the landmarks, from left to right, say, remains the same between views. This accounts for many recognition scenarios, in which places are approached from the same general direction.

To implement this we characterise a place using a {\em landmark distribution descriptor} (LDD), which consists of landmark feature vectors stacked in horizontal position order. Comparison of these descriptors then imposes the constraint of maintaining landmark order alongside matching feature vectors. We find that comparison is best achieved by identifying closest landmark pairs within vertical sections of the panorama, corresponding to subsets of adjacent feature vectors in an LDD (we used 3 sections in the experiments), and summing up distances between the respective feature vectors. We ensure view coverage by requiring sufficient numbers of proposal landmarks within each panoramic section. An example is shown in Figure \ref{fig:showcase}a.

For landmark regions and features we follow the same approach as  S\"{u}nderhauf et al. \cite{Suenderhauf-RSS-15} and use {\em Edge Boxes} \cite{Lawrence-ECCV-14} and convolutional neural network (CNN) features, specifically AlexNet \cite{Krizhevsky-NIPS-2012}, followed by Gaussian random projection \cite{Bingham-KDD-2001} for dimensionality reduction. Our use of panoramic sections to ensure view coverage also mirrors the tiling approach adopted in \cite{McManus-RSS-2014} , although it is important to note that landmark ordering was not used in that work. 

To evaluate the approach we carried out experiments using image pair datasets for places in urban environments. Each dataset consisted of 200 places, with one image pair per place taken from different viewing positions. We used Google Streetview and Bing Streetside images so that image pairs were captured at different times and in different conditions. Results demonstrate that for 10 datasets in 6 different cities, our method performs consistently and significantly better compared with that obtained using the method in \cite{Suenderhauf-RSS-15} and whole image matching using SIFT, GIST and CNN features. For example, as shown in Figure \ref{fig:showcase}b, using 25 region proposals per view, on average over all datasets our method yielded an increase in precision (for 100\% recall) of around 12\% over that using whole image CNN, 20\% over that using the method in \cite{Suenderhauf-RSS-15} and 25\%-30\% over that using whole image SIFT and GIST.

The paper is organised as follows. In the next section we provide an overview of the system and details of the implementation of each component. Details of the datasets, the experiments and an analysis of the results is then provided in Section \ref{sec:experiments}. We conclude with an indication of future work.

\section{System Overview \label{sec:overview}}

In common with other place recognition systems we pose the problem as one of matching different views of the same place taken at different times. Labelled  views are assumed to be held in a reference database and the task is to determine the most likely place associated with a test view captured 'online'. In this work we opt to consider the image-pair version of this framework, in which we have one reference view per place and successful recognition corresponds to matching the test view with the one correct reference view above all others. This contrasts with the majority of other evaluations, which have been based on matching frames within videos taken along the same route and successful recognition then being defined as matching a test frame in one video with a frame from a window of frames in another reference video. We discuss this further in Section \ref{sec:experiments}. 

\begin{figure}[t]
\begin{center}
\begin{tabular}{|ccccc|}\hline &&&&\\
\hspace*{0.015\textwidth}&\includegraphics[width=0.45\textwidth]{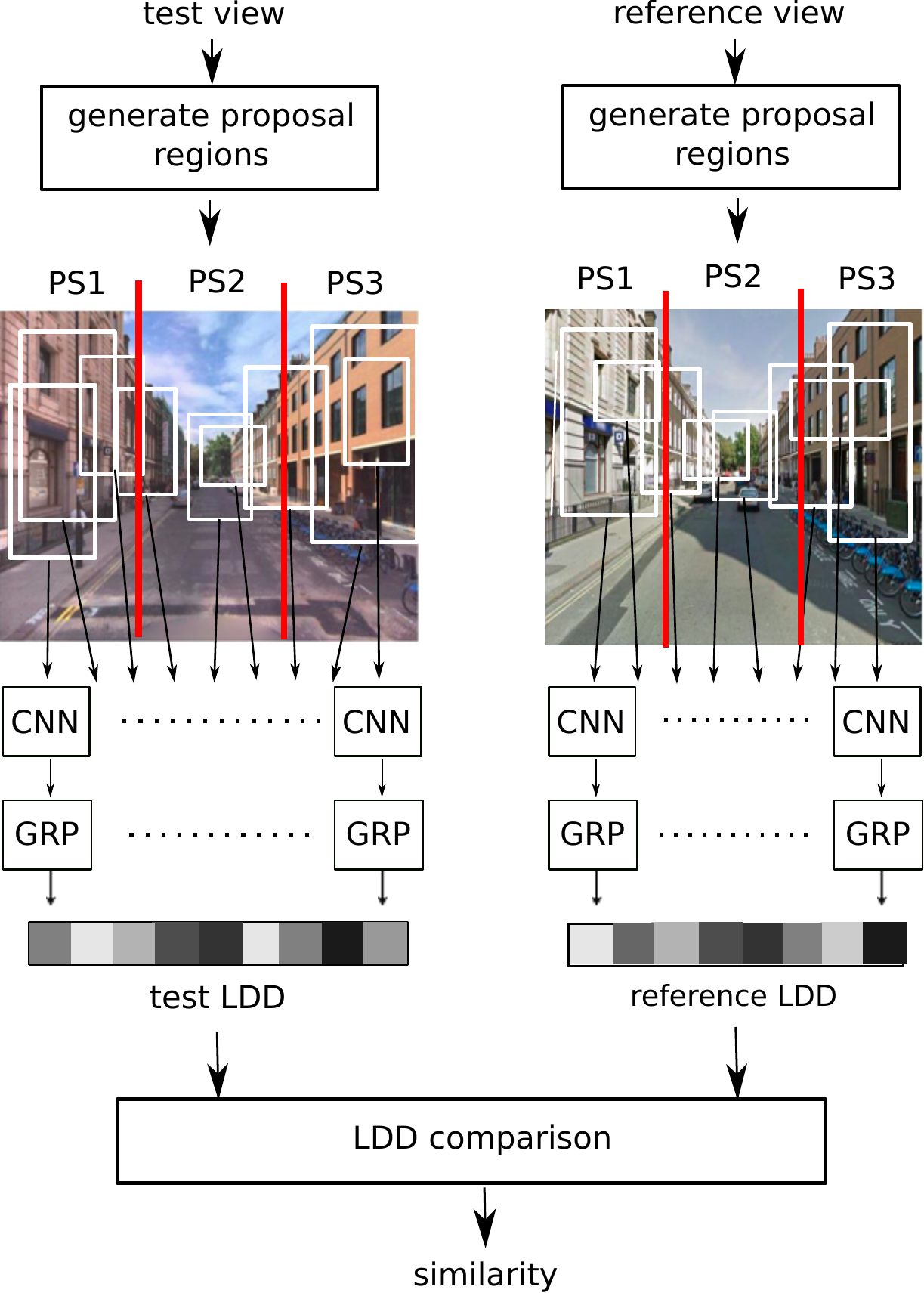}&\hspace*{0.03\textwidth}&
\includegraphics[width=0.35\textwidth]{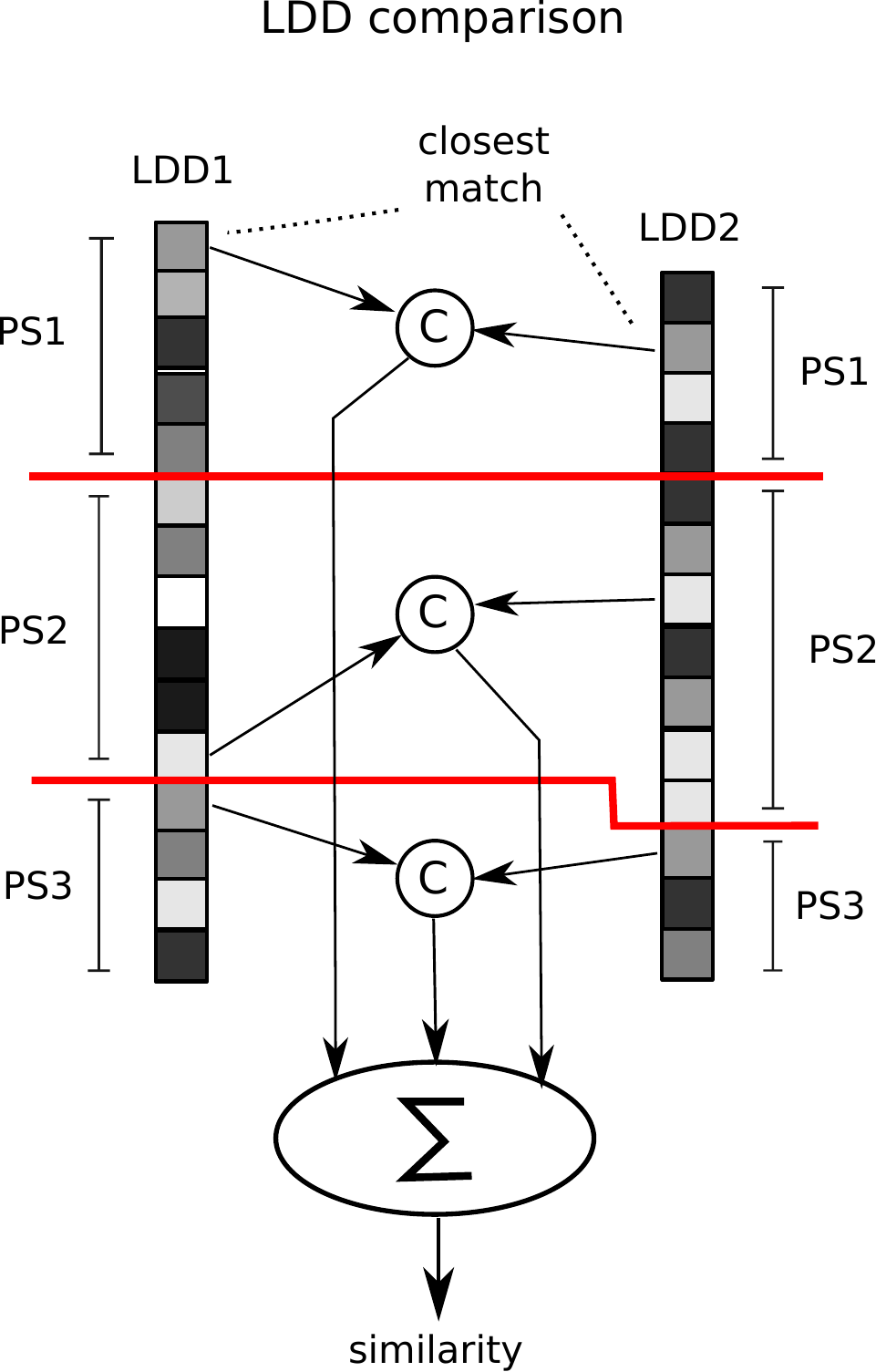}&\hspace*{0.015\textwidth}\\[1ex]
&(a)&&(b)&\\[2ex]\hline
\end{tabular}
\end{center}
\caption{Construction and comparison of landmark distribution descriptors (LDDs). (a) Landmark proposals are generated for the test and reference image using Edge Boxes\cite{Lawrence-ECCV-14}, distributed within panaramic sections PS1-3; landmark features derived from a convolutional neural network (CNN) \cite{Krizhevsky-NIPS-2012} followed by Gaussian random projection (GRP) \cite{Bingham-KDD-2001} are then stacked in horizontal spatial order to form an LDD for each image; descriptors are then compared to derive a distance score between views. (b) Descriptors LDD1 and LDD2 are compared by identifying closest landmark features within each panaramic section and summing the (cosine) distances between them to derive an overall distance score.} 
\label{fig:overview}
\end{figure}

Given the above, we now concentrate on how we match test and reference views. There are two components to this: constructing and comparing LDDs. These  are described below and Figure \ref{fig:overview} provides an illustration of the key elements of each.

\subsection{Constructing LDDs}

There are two main components to constructing an LDD for a given view as illustrated in Fig.\ \ref{fig:overview}a. First, proposal regions are generated, with the aim that a subset of these will correspond to salient landmarks. Second, feature vectors are computed for each of these regions, which are then combined into a single descriptor by stacking them in left-right position order. 

\subsubsection{Landmark proposals}

There are a number of algorithms available for generating proposal regions. In common with S\"{u}nderhauf et al. \cite{Suenderhauf-RSS-15} we choose to use {\em Edge Boxes} as described in \cite{Lawrence-ECCV-14}, which has found widespread use in object recognition and proved to be effective for our application. In brief, a valid edge box is identified as one in which there are a large number of contours wholly enclosed by the box. This is based on the observation that whole contours are likely to correspond to the boundary of distinct objects and hence that such boxes form good proposal regions suitable for further processing. This applies in our case as the landmarks we are interested in such as buildings, windows, trees, etc, satisfy this criterion. Also important is the fact that edge boxes can be found rapidly using fast edge detection combined with fast grouping of pixels into contours. We also make of the edge box ranking in \cite{Lawrence-ECCV-14} in order to limit the number of proposal landmarks and further speed up computation. 

We are also interested in distributing landmark proposals across a view so that we can create a complete description. We do this by partitioning the image vertically and requiring that we select a fix number of the highest ranking landmark proposals in each section. We call these {\em panoramic sections} and in the experiments we used 3 sections: left, middle and right, such as that shown in Fig.\ \ref{fig:overview}a. In the main experiments these were positioned in a regular fashion about the image centre as shown but with overlap between sections to avoid excluding proposals which straddle a section boundary. The alternative is to align the sections according to the content of the view. We experimented with using the vanishing point (VP) as the centre and this proved effective for certain locations. We discuss this further in Section \ref{sec:experiments}. 

More formally, we denote by $L=\{l_1,l_2,\ldots,l_N\}$ the set of landmark proposals in an image discovered by the {\em Edge Boxes} algorithm. We then select a subset of landmarks $\hat{L}$ such that $\hat{L}\subset L$ and

\begin{equation}
\hat{L}=\bigcup_{s=1}^{S} \hat{L}_s
\end{equation}

\noindent where $\hat{L}_s$ is a subset of top ranking proposals in panoramic section $s$ and $S$ is the number of sections, i.e. $S=3$ in the experiments. We fix the number of top ranking proposals according to the section as described in Section \ref{sec:experiments} and deem a proposal to be in a section if its edge box is wholly within the section. Note that when using overlapping sections then individual landmarks can belong to two adjacent sections. This proves to be important when matching landmarks as it reduces the sensitivity to the positioning of section boundaries. 

\subsubsection{Landmark feature vectors}

To match landmarks between views we compute feature vectors to represent the appearance of the regions associated with landmarks. As illustrated in Fig. \ref{fig:overview}a, we again take the same approach as used in \cite{Suenderhauf-RSS-15} and make use of convolutional neural network (CNN) features \cite{Krizhevsky-NIPS-2012} followed by Gaussian random projection (GRP) \cite{Bingham-KDD-2001}  for feature vector size reduction.

CNN features have been shown to provide high levels of invariance to different lighting conditions and viewing positions and hence are ideal for place recognition. Specifically, we used the pre-trained AlexNet network \cite{Krizhevsky-NIPS-2012} as provided by MatConvNet \cite{Vedaldi-ICM-15} and extracted the feature vector of the 3rd convolutional layer ({\em conv3}). Landmark regions were resized to match the required network input size of $227\times 227$ pixels and {\em conv3} produces feature vectors of dimension $13\times 13\times 384=64,896$. 

To reduce the computational load when comparing feature vectors, we project each vector onto a lower dimensional space using GRP \cite{Bingham-KDD-2001}. This is a simple but effective method for dimensionality reduction in which feature vectors are projected onto a significantly smaller number of orthogonal random vectors in such a way that with small error the distances between vectors is maintained. This makes it ideal when matching is based on comparing those distances as in our case. For the experiments we reduced dimensionality down to 1024 for each feature vector without significant impact on performance. In the GRP we used the integer based random projection matrix suggested in \cite{Achlioptas-SPDS-2001}.

For a given view, we construct feature vectors for all the landmark regions in the selected subset of proposals $\hat{L}$ and the vectors corresponding to the section subsets $\hat{L}_s$ then form the LDD for the view. In the experiments we used a total of 25 or 50 proposals per view distributed over 3 panoramic sections and thus each descriptor was of size $25\times 1024$ or $50\times 1024$, respectively.

\subsection{Comparing LDDs}

For place recognition we seek the closest LDD within the database to that of the test image. To compare LDDs we could simply use the Euclidean distance between them. However, this assumes that we have successfully detected the same landmarks in each view, which is unlikely to be the case since we are generating proposals based purely on the appearance of each view individually, and not on the likelihood that a similar landmark exists in the matching view. Hence we transfer the latter constraint into the comparison process. 

As illustrated in Fig.\ \ref{fig:overview}b, we do this by determining the best matching feature vectors (in terms of their cosine similarity) in each of the corresponding panoramic sections. Thus, for example, given two LDDs and using 3 panoramic sections, we seek the best matching pair in each section and then compute an overall matching score corresponding to the sum of the 3 cosine similarities between the feature vectors associated with each pair. We found that using cosine similarity, again in common with \cite{Suenderhauf-RSS-15}, gave improved performance over using a straight Euclidean distance.

More formally, given two descriptors, LDD1 and LDD2, containing landmarks 

\begin{equation}
\{\hat{L}_1^k,\hat{L}_2^k,\ldots,\hat{L}_S^k\}
\end{equation}

\noindent for $k=1,2$, we seek the set of $S$ pairs $(\hat{l}_i^1,\hat{l}_j^2)^s$, $1\leq s\leq S$, such that

\begin{equation}
(\hat{l}_i^1,\hat{l}_j^2)^s=\argmax_{l_i^1\in \hat{L}_s^1, l_j^2\in \hat{L}_s^2} c({\bf v}_i^1,{\bf v}_j^2)
\end{equation}

\noindent where ${\bf v}_i^1$ and ${\bf v}_j^2$ are the feature vectors associated with landmarks $l_i^1$ and $l_j^2$, respectively, and $c({\bf u},{\bf v})={\bf u}.{\bf v}/||{\bf u}||||{\bf v}||$ denotes the cosine similarity between two vectors ${\bf u}$ and ${\bf v}$, where '.' denotes the dot product and $||{\bf u}||$ is the length of ${\bf u}$. To avoid  duplicating matching landmarks, we also require that no landmark can be in more than one matching pair. The overall similarity score between the two LDDs, and hence the two views, is then given by the sum of the $S$ cosine similarities, i.e.

\begin{equation}
sim_{12}=\sum_{\substack{(\hat{l}_i^1,\hat{l}_j^2)^s\\1\leq s\leq S}} c(\hat{{\bf v}}_i^1,\hat{{\bf v}}_j^2)
\end{equation}

\section{Experiments \label{sec:experiments}}

\begin{figure}[t]
\begin{center}
\begin{tabular}{ccc}
\includegraphics[width=0.3\textwidth]{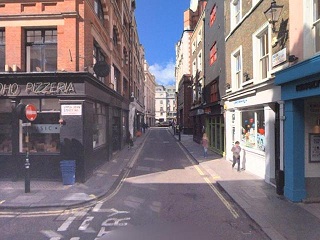}&
\includegraphics[width=0.3\textwidth]{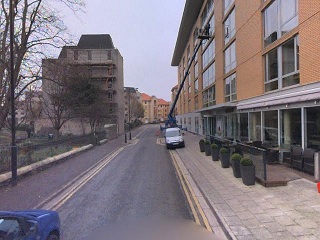}&
\includegraphics[width=0.3\textwidth]{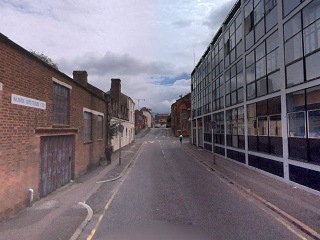}\\
\includegraphics[width=0.3\textwidth]{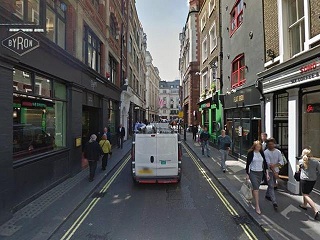}&
\includegraphics[width=0.3\textwidth]{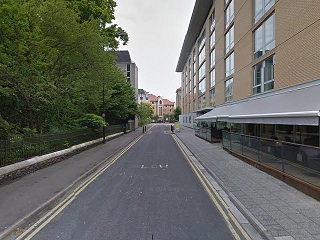}&
\includegraphics[width=0.3\textwidth]{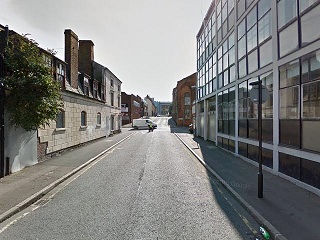}\\[3ex]
\includegraphics[width=0.3\textwidth]{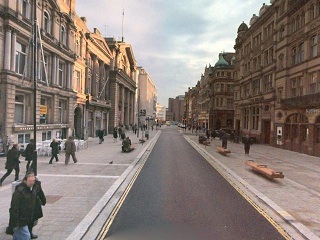}&
\includegraphics[width=0.3\textwidth]{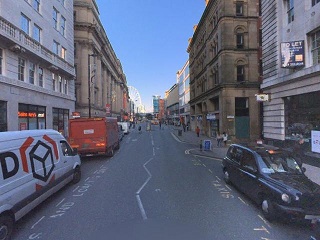}&
\includegraphics[width=0.3\textwidth]{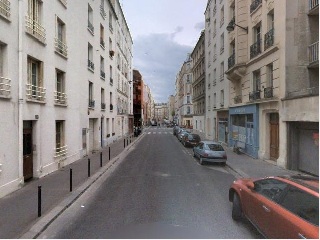}\\
\includegraphics[width=0.3\textwidth]{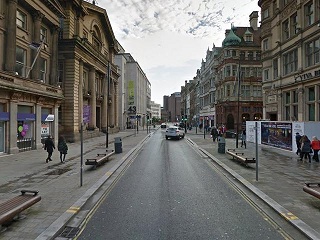}&
\includegraphics[width=0.3\textwidth]{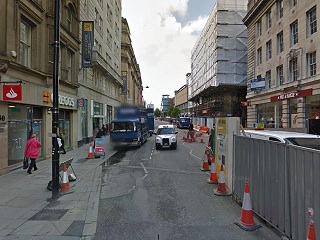}&
\includegraphics[width=0.3\textwidth]{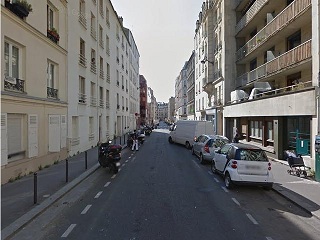}
\end{tabular}
\end{center}
\vspace*{-1ex}
\caption{Examples of view pairs from each of the 6 cities in the 10 datasets used in the experiments. The pairs are shown one above the other and there are 3 pairs per row.}
\label{fig:example-city-pairs}
\vspace*{-1ex}
\end{figure}

\subsection{Datasets}

We evaluated our method using multiple image pair datasets taken from urban environments. Our motivation for using image-pairs in contrast to matching frames in video sequences as used by others is twofold. First, we believe that it presents a more challenging test, since recognition is based on matching with only one alternative view as opposed to matching to one of multiple frames in a video (corresponding to the vicinity of a place). Secondly, it means that we can easily create large datasets corresponding to random places using the images taken from online mapping services such as Google Streetview, Bing Streetside and Mapillary. Using images from more than one of these also enables us to evaluate performance under different viewing conditions.

For the experiments reported here we used datasets obtained from Google Streetview and Bing Streetside. Specifically, we selected 200 random locations in 6 different  cities - London, Bristol, Birmingham, Liverpool, Manchester and Paris - and for each location we selected images taken in roughly the same sort of direction but displaced by between approximately 5 and 10 meters. We selected one image from Streetview and one from Streetside for each location. This is ideal as the images were taken at different times and under different lighting and visibility conditions. We used datasets from different cities to enable us to evaluate the performance of the method for differing types and characteristics of architecture and urban layout. We tested the method on 10 datasets in all, using 3 from London and 3 from Bristol in order to test for any variation in performance within the same city. In total, the evaluation involved 2,000 different locations.

Example image pairs from the different cities are shown in Fig.\ \ref{fig:example-city-pairs}. Note that although the physical distance between the viewing positions is not great, there is a significant change in structural appearance which when coupled with the differences caused by different light and visibility conditions makes recognition far from straightforward. Notable difficulties include the presence of pedestrians and vehicles, significant changes in scale of buildings, some buildings disappearing from view whilst others come into sharper focus, and so on. However, careful observation should reveal that distribution of the key visible landmarks is maintained across the two views. It is this characteristic that we aim to exploit in this work.

\subsection{Comparison methods}

We compared the performance of our method with four other methods: the CNN landmark matching method of S\"{u}nderhauf et al. \cite{Suenderhauf-RSS-15}; whole image CNN matching \cite{Suenderhauf-IROS-15}; whole image SIFT matching; and whole image GIST matching. Relevant details for each method are given below.

\begin{description}
\item{CNN Landmark matching (CLM)}

As noted earlier, the primary difference between our method and that in \cite{Suenderhauf-RSS-15} is that matching in the latter is based on finding similar landmarks across both views, irrespective of relative position. Specifically, best matching pairs of CNN-GRP feature vectors are selected from edge box proposals based on cosine similarity and the overall similarity between two views is then the sum of the cosine similarities, weighted by a measure of similarity in box size. For comparison, we evaluated two versions of this method, one using 25 (proposals CLM-25) and one using 50 proposals( CLM-50)\footnote{For clarity with respect to our experiments, we should note that we found that the similarity metric provided in \cite{Suenderhauf-RSS-15} did not give good performance and so in the interests of fairness we used a modified version which gave significantly better performance. Specifically, we modified equations (2) and (3) in \cite{Suenderhauf-RSS-15} to be $s_{ij} = 1 - (\frac{1}{2} (\frac{|w_{i} - w_{j}|}{max(w_{i},w_{j})} + \frac{|h_{i} - h_{j}|}{max(h_{i},h_{j})}))$ and $S_{ab} = \frac{1}{n_{a} \cdot n_{b}} \sum_{ij} (d_{ij} \cdot s_{ij}))$, respectively.}

\item{CNN matching (CWI)}

In this method, we used the same CNN-GRP features vectors as in \cite{Suenderhauf-RSS-15} and in our method, but comparison between views was based on a single feature vector computed for the whole image. Cosine similarity was again used as the comparison metric. The method is similar to that used in \cite{Suenderhauf-IROS-15}.

\item{Dense SIFT matching (SIFT)}

For this method we used a dense keypoint version of matching SIFT descriptors across both views \cite{Lowe-IJCV-2004}. Specifically, we used the implementation as provided in the VLFeat library \cite{Vedaldi-ICM-10}.

\item{GIST matching (GIST)}

Finally, we compared our method with whole image GIST matching, based on the implementation provided by Oliva and Torralba as described in \cite{Oliva-IJCV-2001}.

\end{description}

\subsection{Results}

We compared the performance of our method against that of the comparison methods for all 10 datasets. Each dataset contained 200 view pairs from different locations, with one view taken from Streetview and the other from Steetside. In each evaluation, we used all the Streetside images as test images and the Streetview images were used as the reference images.  We used precision ($P$) and recall ($R$) to measure performance, using the following definitions of $P$ and $R$

\begin{equation}
P=\frac{tp}{tp+fp} \hspace{5em} R=\frac{tp}{tp+fn}
\end{equation}

\noindent where $tp$, $fp$ and $fn$ denote the number of true positives, false positives and false negatives, respectively. A true positive was recorded if the test image was matched with the reference image taken at the same location, a false positive was recorded if the test image was matched with a reference image taken at a different location, and a false negative was recorded if a test image was deeemed not to match any of the reference images based on a threshold of the ratio between the closest and second closest matches. Variation of this threshold also enabled us to create precision-recall curves as given below. Note that our datasets do not contain any true negatives.

We evaluated two versions of our method, one using 25 landmark proposals and one using 50 landmark proposals. In each case we used 3 panoramic sections, with 50\% overlap between sections. The image sizes were $640\times 480$ pixels for both Streetview and Streetside and we used sections of size $320$ pixels. We fixed the number of top ranked proposals selected from each section to be (5,15,5) when using 25 proposals (in left to right order) and (10,30,10) when using 50 proposals. The larger number of proposals in the central section proved to have a significant impact on performance.

\setlength{\tabcolsep}{4pt}
\begin{table}[t]
\centering
\caption{Recorded precision values for 100\% recall for all 10 datasets using all 7 comparison methods.}
\vspace{2ex}
\label{tab:precision}
\begin{tabular}{lccccccc}
\hline \noalign{\smallskip}
 & \textbf{LDD-25} & \textbf{CLM-25} &  \textbf{LDD-50} & \textbf{CLM-50}  & \textbf{SIFT} & \textbf{GIST} & \textbf{CWI} \\ \hline \noalign{\smallskip}
London1 & {\bf 84.5} & 55 &  {\bf 88} & 73.5 &  59.5 & 47 & 66 \\
London2 & 83 & 68 &  {\bf 90} & {\bf 84}  & 58 & 44 & 74 \\
London3 & {\bf 72} & 57 & {\bf 83} & 69  & 51 & 58 & 64 \\
Bristol1 & {\bf 66.5} & 51.5 & {\bf 68.5} & 58 & 51.5 & 33 & 60.5 \\
Bristol2 & {\bf 63.5} & 50.5  & {\bf 65.5} & 59.5  & 40.5 & 26 & 54.5 \\
Bristol3 & 59.5 & 47 & {\bf 67} & {\bf 64.5} & 48 & 37 & 61 \\
Birmingham & {\bf 62} & 44 & {\bf 71.5} & 60 &  26.5 & 38 & 44 \\
Manchester & {\bf 69} & 50.5 & {\bf 71.5} & 63.5 & 33.5 & 33.5 & 63 \\
Liverpool & {\bf 74} & 46 & {\bf 75} & 62  & 52.5 & 40.5 & 53 \\
Paris & {\bf 61} & 35  & {\bf 70.5} & 49  & 40 & 35 & 38 \\
\hline \noalign{\smallskip}
\textbf{Avg} & \textbf{69.5} & \textbf{50.45} &  \textbf{75.05} & \textbf{64.3} & \textbf{46.1} & \textbf{39.2} & \textbf{57.8} \\ \hline \noalign{\smallskip}
\end{tabular} 
\end{table}

Table \ref{tab:precision} shows the precision values recorded for the different methods at 100\% recall, i.e. so that all matches are accepted as positives. Note that for all datasets the LDD-50 method gives the best performance and apart from two datasets, the LDD-25 method gives the next best results. The latter is significant since the computational load is halved when using 25 proposal landmarks (the bottle neck is the computation of the CNN feature vectors)  and thus it is interesting to note that good performance is still maintained from our method using the smaller number of proposals. This contrasts with the CLM method which performs significantly worse when using only 25 proposals and notably worse than using whole image CNN. We believe that this is a direct result of our method using the sptial distribution of the landmarks which provides a key characteristic to distinguish between views. 

\begin{figure}[t]
\begin{center}
\begin{tabular}{ccc}
\includegraphics[width=0.45\textwidth]{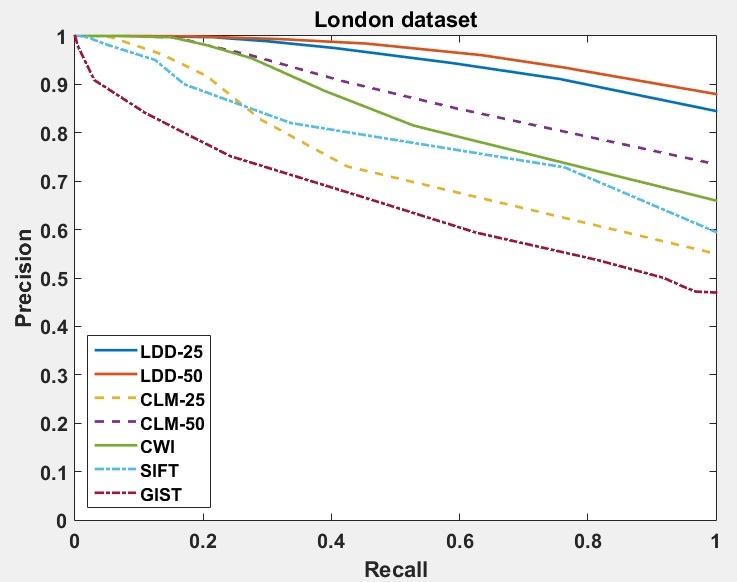}&&
\includegraphics[width=0.45\textwidth]{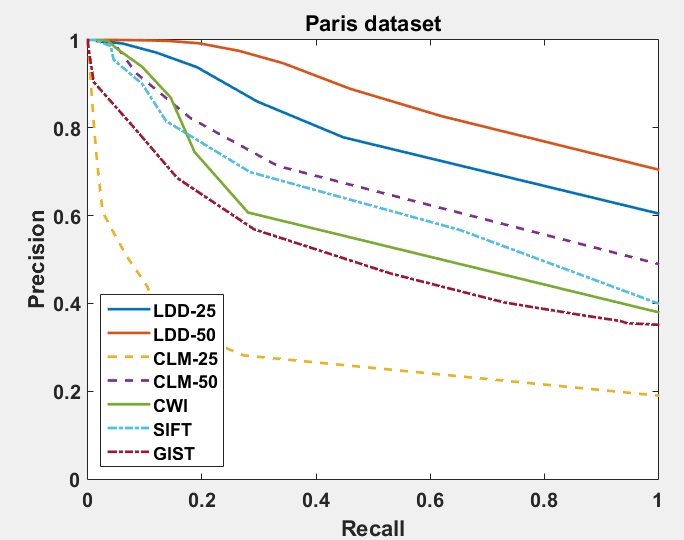}\\[1ex]
(a)&&(b)
\end{tabular}
\end{center}
\vspace*{-3ex}
\caption{Precision recall curves obtained for all comparison methods for (a) the London1 dataset and (b) the Paris dataset. } 
\label{fig:prcurves}
\end{figure}


\begin{figure}[h]
\begin{center}
\begin{tabular}{cccc}
\includegraphics[width=0.23\textwidth]{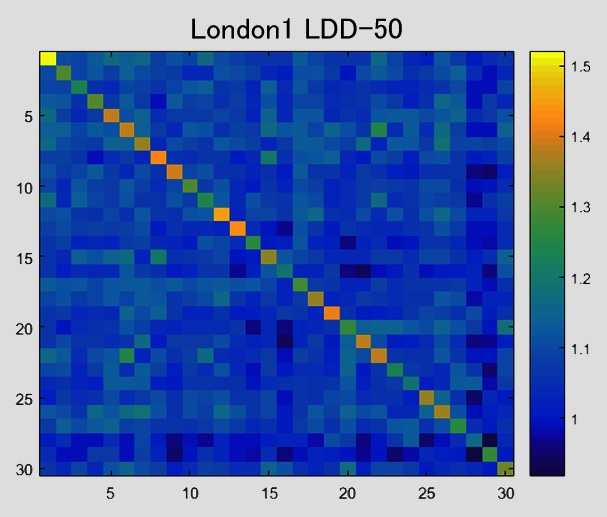}&
\includegraphics[width=0.23\textwidth]{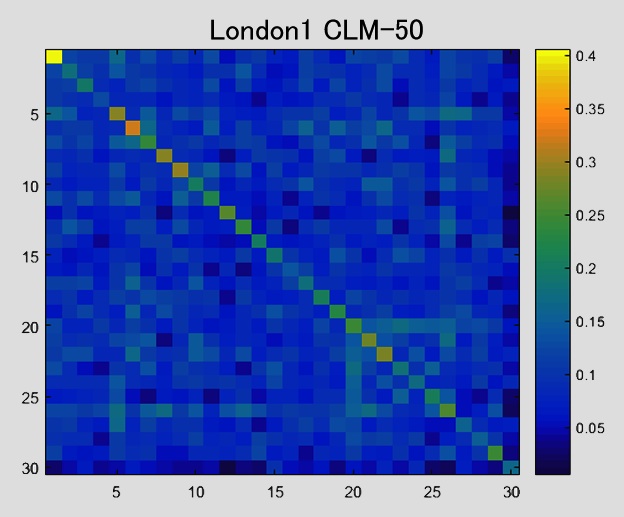}&
\includegraphics[width=0.23\textwidth]{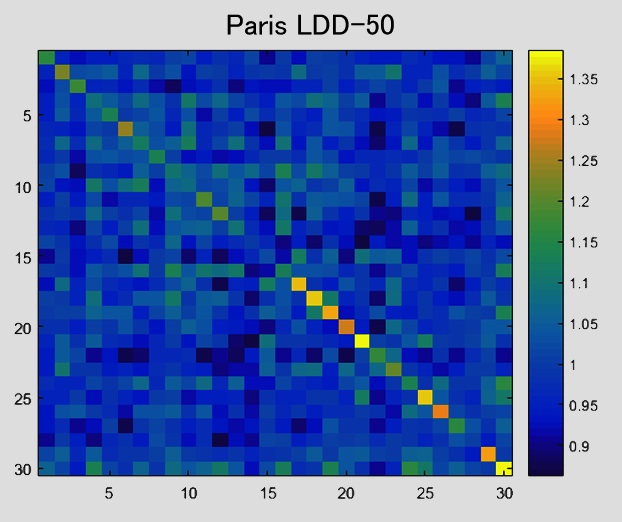}&
\includegraphics[width=0.23\textwidth]{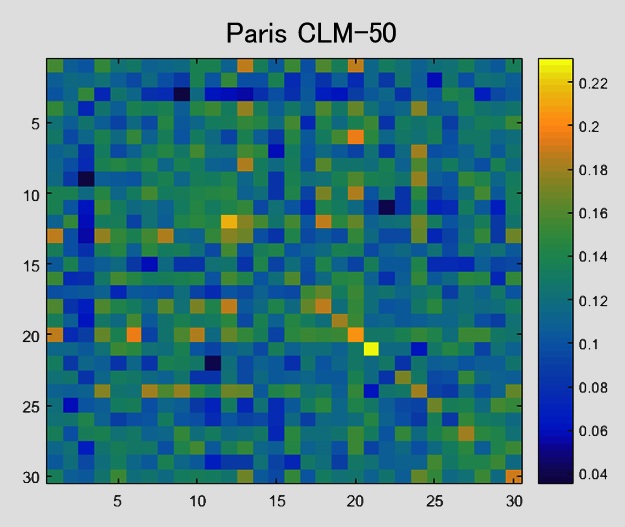}\\[1ex]
(a)&(b)&(c)&(d)
\end{tabular}
\end{center}
\vspace*{-3ex}
\caption{Confusion matrices showing recorded similarity scores for 30 locations in the London1 and Paris datasets using (a)-(b) LDD-50  and (c)-(d) CLM.} 
\label{fig:confuse-matrix}
\end{figure}

\begin{figure}[t]
\begin{center}
\begin{tabular}{ccc}
\includegraphics[width=0.3\textwidth]{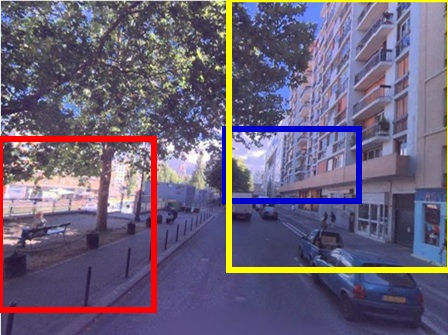}&
\includegraphics[width=0.3\textwidth]{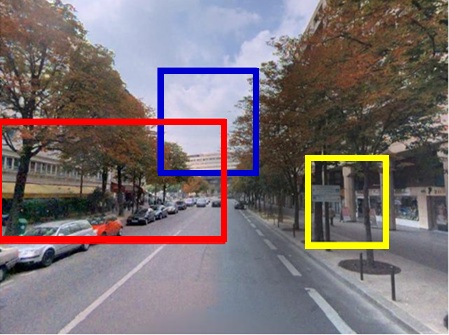}&
\includegraphics[width=0.3\textwidth]{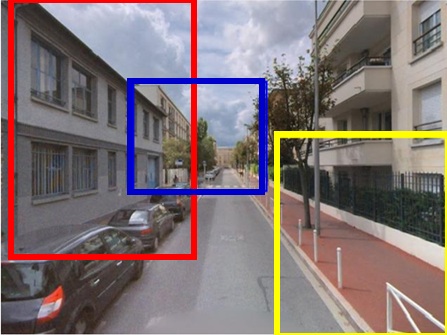}\\
\includegraphics[width=0.3\textwidth]{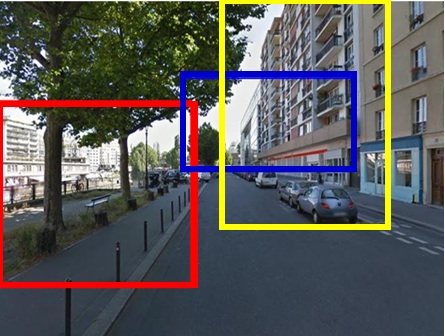}&
\includegraphics[width=0.3\textwidth]{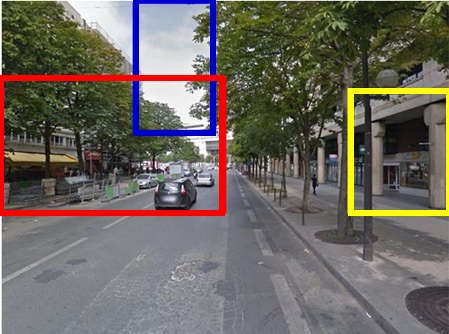}&
\includegraphics[width=0.3\textwidth]{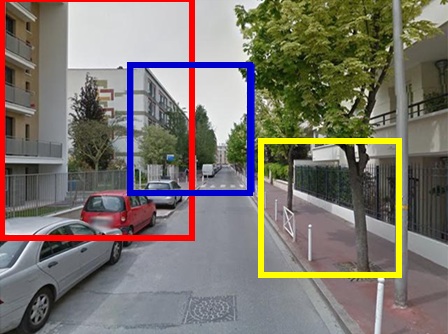}\\[2ex]
\includegraphics[width=0.3\textwidth]{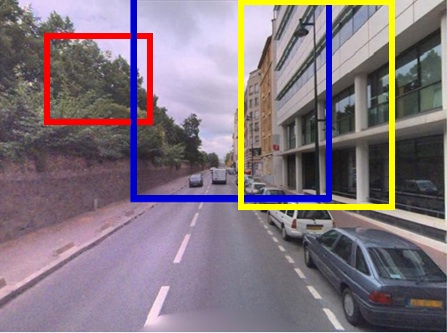}&
\includegraphics[width=0.3\textwidth]{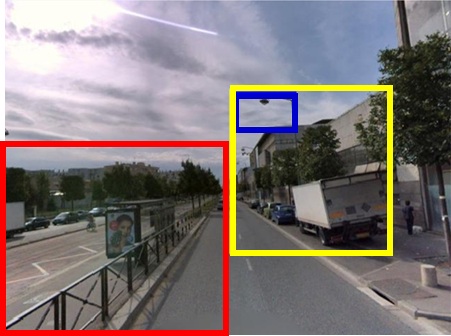}&
\includegraphics[width=0.3\textwidth]{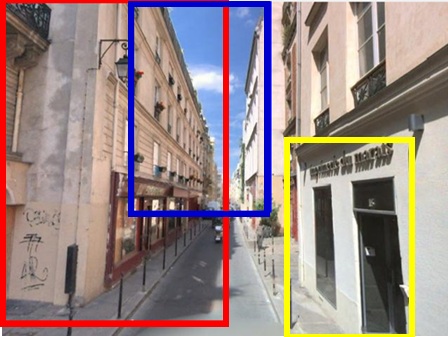}\\
\includegraphics[width=0.3\textwidth]{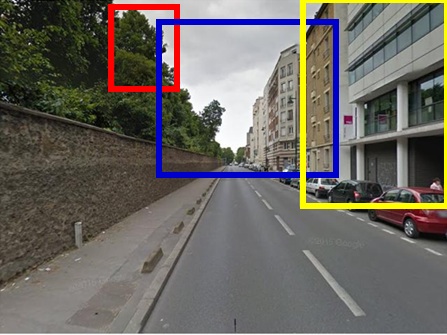}&
\includegraphics[width=0.3\textwidth]{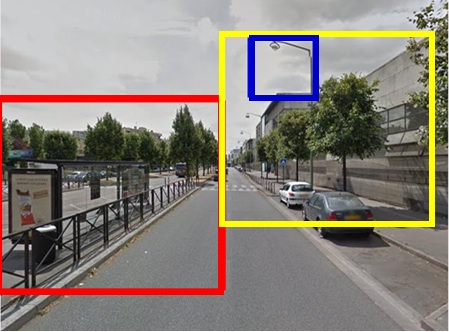}&
\includegraphics[width=0.3\textwidth]{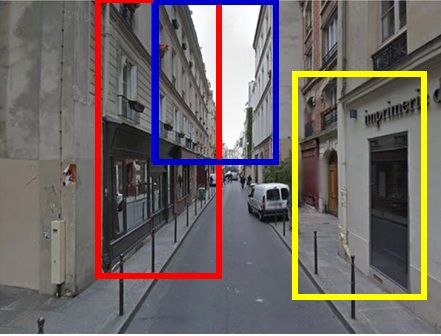}\\[2ex]
\end{tabular}
\end{center}
\vspace*{-3ex}
\caption{Examples of correct view matches obtained using the LDD-50 method. Matches are shown one above the other and there are 3 matches per row.}
\label{fig:correct1}
\vspace*{-1ex}
\end{figure}

Figure \ref{fig:prcurves} shows the variation in precision as we reduce recall by increasing the number of false negatives via the threshold on the ratio of the closest and second closest matches for the two datasets London1 and Paris. Note that in both cases both versions of our method LDD-25 and LDD-50 outperform the other methods. Again, the difference in LDD-25 and CLM-25 is noticable, with the former achieving almost a 30\% gain in precision, corresponding to correct recognition of over 60 places compared with the latter, using the same number of proposal landmarks. This illustrates clearly the advantage of using landmark distribution to characterise views. To illustrate the distinguishing power of our method, Fig.\ \ref{fig:confuse-matrix} shows confusion matrices for the same two datasets using methods LDD-50 and CLM-50, where we have used 30 randomly selected location pairs rather than all 200 to aid clarity. These show the similarity scores between test and reference views. Note the high values down the main diagonal for the LDD-50 method indicating strong distinction of the correct places and contrast this with the closeness of the values obtained using CLM-50 method, especially for the Paris dataset.

\begin{figure}[t]
\begin{center}
\begin{tabular}{ccc}
\includegraphics[width=0.3\textwidth]{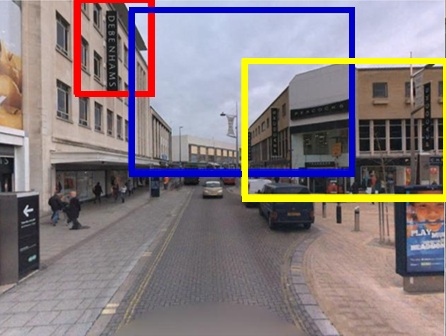}&
\includegraphics[width=0.3\textwidth]{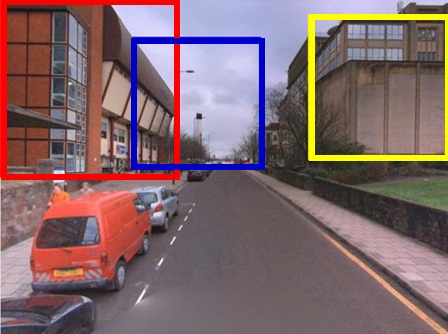}&
\includegraphics[width=0.3\textwidth]{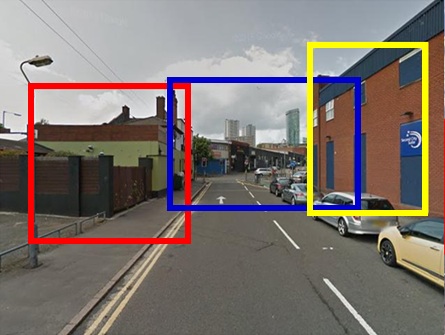}\\
\includegraphics[width=0.3\textwidth]{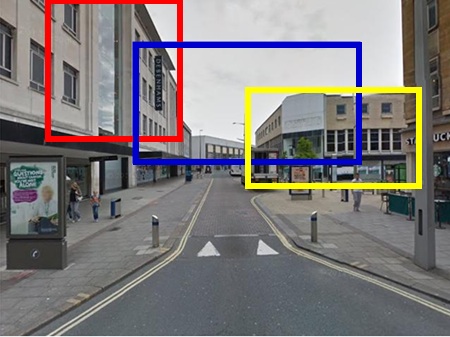}&
\includegraphics[width=0.3\textwidth]{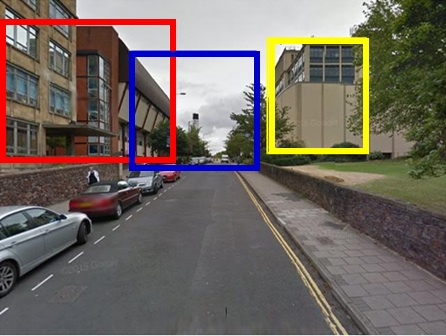}&
\includegraphics[width=0.3\textwidth]{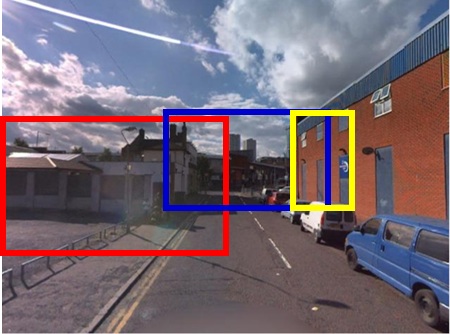}
\end{tabular}
\end{center}
\vspace*{-2ex}
\caption{Further examples of correct view matches obtained using the LDD-50 method. }
\label{fig:correct2}
\end{figure}

To illustrate the landmarks that are being found by our method to enable correct recognition of places, Figures \ref{fig:correct1} and \ref{fig:correct2} shows examples of views which have been correctly matched. {\em None of these examples were correctly matched by the other methods}. In each case, the best matching landmarks found in each panoranic section are shown in colour, where the colours indicate corresponding landmarks in each view. Pairs are shown above one another and each row shows 3 pairs. Note the difference in appearance and structure between the views, especially the changes in vegetation and building structure, but also note that with careful observation they can be seen to be the same places. These are challenging examples and it is encouraging that our method is able to correctly match the views.

\begin{figure}[t]
\begin{center}
\begin{tabular}{cc}
\includegraphics[width=0.3\textwidth]{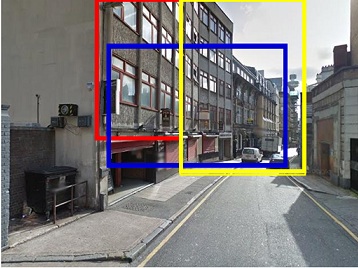}&
\includegraphics[width=0.3\textwidth]{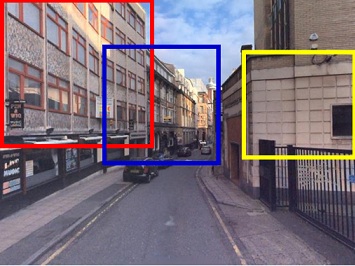}\\[2ex]
\includegraphics[width=0.3\textwidth]{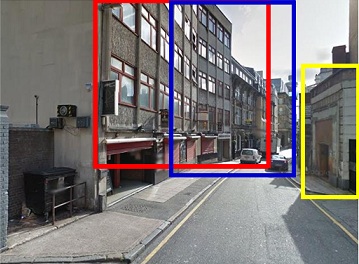}&
\includegraphics[width=0.3\textwidth]{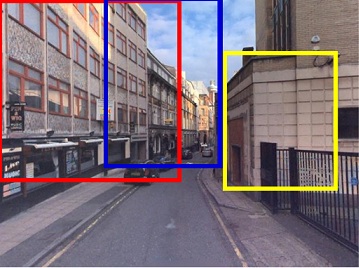}
\end{tabular}
\end{center}
\vspace*{-2ex}
\caption{Use of the view vanishing point to center panoramic sections improves matching of landmarks (bottom) over that obtained using the image center (top).}
\label{fig:vpexample}
\vspace*{4ex}
\end{figure}

\begin{figure}[!ht]
\begin{center}
\begin{tabular}{ccc}
\includegraphics[width=0.3\textwidth]{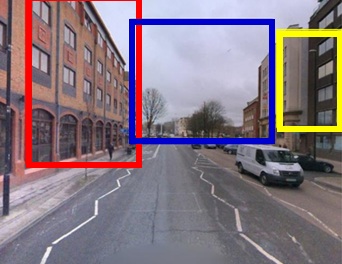}&
\includegraphics[width=0.3\textwidth]{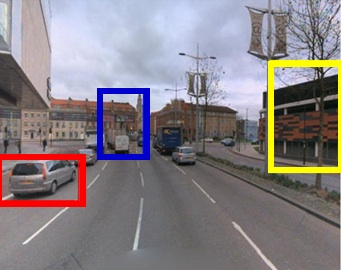}&
\includegraphics[width=0.3\textwidth]{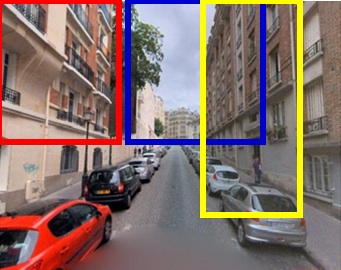}\\
\includegraphics[width=0.3\textwidth]{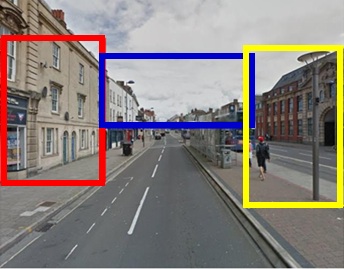}&
\includegraphics[width=0.3\textwidth]{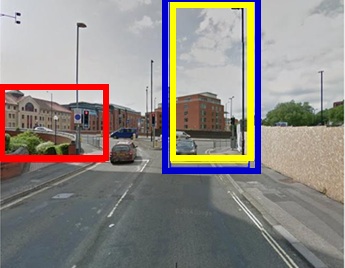}&
\includegraphics[width=0.3\textwidth]{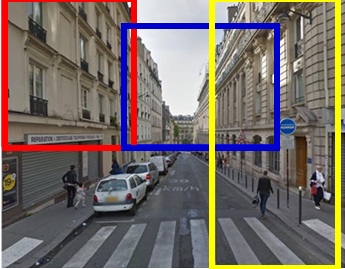}\\
\includegraphics[width=0.3\textwidth]{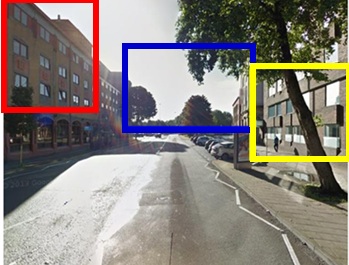}&
\includegraphics[width=0.3\textwidth]{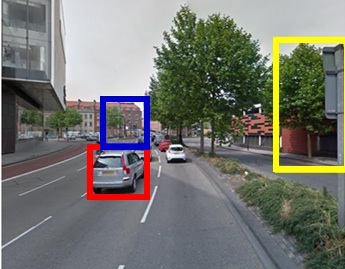}&
\includegraphics[width=0.3\textwidth]{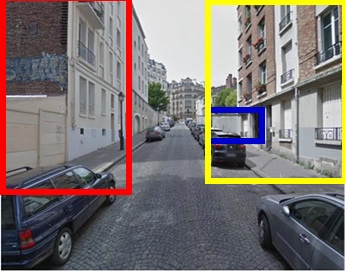}\\
\end{tabular}
\end{center}
\vspace*{-2ex}
\caption{Examples of incorrectly matched views obtained using the LDD-50 method.}
\label{fig:incorrect}
\end{figure}

We also experimented with adapting the positioning of the panoramic sections according to view content rather than simply dividing up the image evenly into 3 sections about the image centre. Instead, we computed the location of the vanishing point in each view, using the method described in \cite{Kong-CVPR-2009}, and if within the image, we used this to centre the middle section, with appropriate adaptation of the two outer sections. In many cases this had little impact since the VP was often close to the image centre. However in a number of cases it did make a difference and resulted in correct matching of places which were previously incorrectly matched. An example is shown in Fig.\ \ref{fig:vpexample}. The top row shows a pair of views of the same place with selected landmark regions derived using panoramic sections centred about the image center. This proved not to be the best match for the left hand test image and hence resulted in an incorrect match. Clearly the detected landmarks in each view do not correspond to the same landmarks in the  scene. In contrast, shifting the sections to the right in both views after detecting the VP in each, results in correspondence between the detected landmarks and this resulted in a successful match. Although encouraging, these are only provisional results and further work is needed to determine the generality of using the VP in this way.

Finally, Fig.\ \ref{fig:incorrect} shows 3 examples in which our method fails to match the correct view. The top row shows the test images, the middle row shows the incorrectly matched view and the bottom row the correct view. Note that these are particularly challenging examples and are further complicated by landmarks being detected on vehicles which are not present in both views. How to deal with cases such as these will be the subject of further research.


\section{Conclusions and future work}

We have presented a new method for visual place recognition based on matching landmark regions represented by CNN features. The key contribution is the encoding of relative spatial position of the landmarks via the use of the landmark distribution descriptors (LDD). Although the method has aspects in common with the CLM method of  S\"{u}nderhauf et al. \cite{Suenderhauf-RSS-15}, we have demonstrated that the use of LDDs has a major impact on performance, with significant gains in precision, not only over CLM but also over the other whole image techniques. It is important to point out that the gains in precision amount to significant gains in the numbers of correctly recognised places, with a 20\% gain corresponding to 40 locations. 

In the future we intend to investigate the performance of the method using different datasets, including video. We will also investigate further the benefits of using  VPs to better position the panoramic sections. Also of interest is the potential for extending the idea of landmark distribution matching to more general cases in which landmark positioning changes due to changes in viewpoint. As the two are linked through geometry and motivated by the ideas and method described in \cite{Frampton-BMVC-2013}, it may be possible to build this into a contraint for matching views which are widely disparate.


\bibliographystyle{splncs}
\bibliography{arXiv.bib}

\end{document}